\documentclass[pdflatex,sn-basic]{sn-jnl}

\usepackage{graphicx}%
\usepackage{multirow}%
\usepackage{amsmath,amssymb,amsfonts}%
\usepackage{amsthm}%
\usepackage{mathrsfs}%
\usepackage[title]{appendix}%
\usepackage{xcolor}%
\usepackage{textcomp}%
\usepackage{manyfoot}%
\usepackage{booktabs}%
\usepackage{algorithm}%
\usepackage{algorithmicx}%
\usepackage{algpseudocode}%
\usepackage{listings}%

\usepackage{makecell}

\theoremstyle{thmstyleone}%
\theoremstyle{thmstyletwo}%

\theoremstyle{thmstylethree}%

\makeatletter
\gdef\orcidlogo{} 
\makeatother

\begin{document}

\title[Article Title]{Class-Aware Reinforcement Learning for Counterfactual Explanation Generation}


\author[1]{\fnm{Muhammad Adil} \sur{Saleem}}\email{adilsaleem@iba.edu.pk}
\equalcont{These authors contributed equally to this work.}

\author*[1]{\fnm{Syed Ali} \sur{Raza}}\email{saliraza@iba.edu.pk}
\equalcont{These authors contributed equally to this work.}

\author[2,3,4]{\fnm{Mary-Anne} \sur{Williams}}\email{Mary-Anne.Williams@cba.com.au}

\affil*[1]{\orgdiv{Department of Computer Science}, \orgname{Institute of Business Administration Karachi}, \orgaddress{\street{University Road}, \city{Karachi}, \postcode{75270}, \country{Pakistan}}}

\affil[2]{\orgname{Commonwealth Bank of Australia}, \orgaddress{\street{The Foundry, 1 Locomotive St, Eveleigh NSW 2015}, \city{Sydney}, \postcode{2015}, \state{NSW}, \country{Australia}}}

\affil[3]{\orgname{Stanford University}, \orgaddress{\city{Stanford}, \postcode{94305}, \state{California}, \country{USA}}}

\affil[4]{\orgname{University of New South Wales}, \orgaddress{\city{Sydney}, \postcode{2052}, \state{NSW}, \country{Australia}}}


\abstract{Counterfactual explanations (CFEs) enhance the interpretability of black-box models by generating alternative instances with adjusted feature values that achieve a contrastive outcome. Reinforcement learning (RL) offers a promising approach for CFE generation, enabling efficient exploration of counterfactual instances while ensuring control over key metrics like validity, sparsity, and proximity. Previous studies have formulated RL states exclusively using features derived from the predictors in the supervised dataset. This study explores the impact of including an instance's predicted class, alongside features derived from the predictors, in the RL state representation for generating CFEs. The hypothesis is that class-awareness enhances exploration efficiency and improves policy optimality.
We compare the proposed class-aware RL method with the class-blind RL method, which is similar but excludes the instance's class information from the state representation. The comparison was conducted using seven datasets from diverse domains, varying in size. The results show that during training, class-aware RL offers benefits in terms of convergence speed, reward optimization, and episode length reduction. Moreover, it generates significantly more valid CFEs compared to class-blind RL. Finally, the instance's class-based feature consistently ranks among the most influential predictors in RL's action-selection, as shown by the SHAP and LIME values, underscoring the significance of class-awareness in RL for CFE generation.
The impact is heightened clarity, faster learning, improved validity, and more effective counterfactual generation across diverse datasets.
}




\maketitle

\section{Introduction}\label{sec1}

Understanding the decision-making processes of machine learning models has grown increasingly important, giving rise to the field of Explainable AI (XAI), which lacks a standardized definition for explanation \cite{williamsExplainableArtificialIntelligence2021}. As the name suggests, XAI focuses on explaining how these models reach their decisions. XAI enhances the transparency and trustworthiness of machine learning (ML) models, which is crucial for their adoption in sensitive applications such as healthcare \cite{abbas2025explainable} and finance \cite{arsenault2025survey}.
One effective approach within XAI involves assessing feature importance, which identifies the significance of individual features in the model's decision-making process. 
Notable techniques employed for gauging feature importance include SHAP \cite{Lundberg2017Unified} and LIME \cite{Ribeiro2016"Why}. Both SHAP and LIME are post-hoc (i.e., applied after model training), model-agnostic (i.e., independent of the underlying model type).

Counterfactual Explanation (CFE) \cite{wachter2017} offers a different form of feature importance-based explanation within the context of XAI.
Unlike explicit feature importance, CFEs generate instances similar to the original but with different outcomes \cite{Molnar2022Interpretable}. The differences between the features of a counterfactual instance and those of the original instance offer implicit explainability, shedding light on the key factors driving a black-box model's predictions. CFE offers critically important advantages such as actionability, empowering users to modify the outcome with minimal change. The term "minimal change" varies based on the target measure, such as sparsity or proximity, resulting in different counterfactual explanations (CFEs). 
CFEs have been applied across diverse domains, including finance, customer analytics, and cybersecurity, motivating evaluation across multiple datasets \cite{kalasampath2025literature, grath2018interpretable,li2023counterfactual,cumi2024counterfactual}.

DiCE leverages gradient-based optimization to generate sets of diverse counterfactual examples \cite{mothilal2020explaining}. However, its reliance on gradients makes it less effective for non-differentiable models or true black-box models, where gradient information is inaccessible. In contrast, NICE can handle non-differentiable models but relies on the nearest unlike neighbor to generate CFEs \cite{brughmans2024nice}. This reliance restricts the search space for counterfactuals and may overlook optimal explanations located in other regions of the feature space.

Modeling CFE generation as a search problem, transitioning from an instance to the desired CFE, involves adjusting features until the CFE is obtained. Reinforcement learning (RL) effectively models this process, efficiently navigating feature space through a learned policy.
Unlike instance-wise optimization-based CFE methods (e.g., Wachter et al. \cite{wachter2017}), which solve a separate optimization problem for each input, RL frames CFE generation as a sequential decision-making process and learn an amortized policy that can be reused across samples. In contrast to generator-based CFE approaches such as CounterNet \cite{guo2023counternet} and VCNet \cite{guyomard2022vcnet}, our RL-based formulation operates in a fully black-box setting and applies constrained feature modifications incrementally.
As demonstrated by Chen et al. \cite{Chen2022ReLAX:}, RL excels in generating CFEs on benchmark datasets, showing promising performance across sparsity, validity, and proximity metrics. Similar work was proposed by \cite{verma2022amortized} suggesting offline policy training, labeled \textit{amortized inference}, for faster CFE generation.  Although RL has proven successful for CFE generation, existing approaches have not explored the inclusion of the instance?s class in the state representation of the RL model. Intuitively, incorporating the instance's predicted class is important because CFE modifies the instance until its actual class changes. Contrary to intuition, including the predicted class information in machine learning is seen as data leakage, which may explain why researchers have not pursued this approach.

\begin{figure}[htbp]
	\centering
	\includegraphics[width=1.0\linewidth]{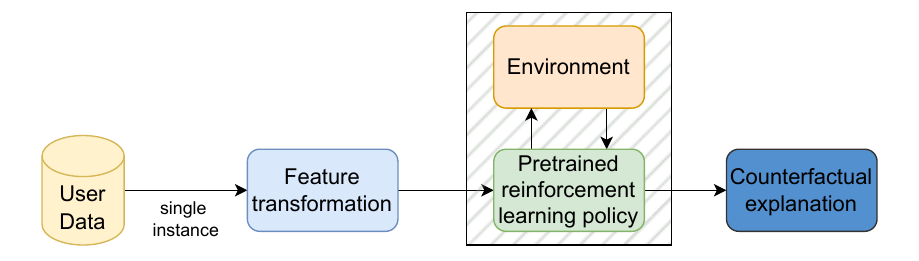}
	\caption{Illustration of the process for generating counterfactual explanations using reinforcement Learning. }
	\label{fig:RL-CFE-generation}
\end{figure}

In this paper, we propose an RL-based approach for CFE generation, enhancing the state representation by including a predicted class-based feature of the instance along with features derived from the predictors in supervised tabular data.
The inclusion of instance's predicted class, readily available as the model's prediction, was expected to achieve higher terminal reward which is indicative of obtaining a better CFE. The motivation behind introducing information about the instance's predicted class was to provide the RL model with additional context. This context aids in navigating unexplored states by enabling the model to make informed decisions, prioritizing actions that lead to a different class when the optimal action in a given state is uncertain. 
The task of generating CFEs is framed as a Markov Decision Process (MDP), with the components of states, actions, and rewards formally defined. An environment is designed for each dataset to facilitate agent-environment interaction. The CFE generation process is treated as an episodic task, starting from a state initialized with the original instance's feature values. At each step, the RL agent adjusts one of the mutable features, ultimately identifying feature values that switch the class or fail to do so within a predefined number of steps. Rewards are carefully designed to balance key CFE metrics, including validity, sparsity, and proximity.

A high-level overview of our methodology for generating CFEs is shown in Figure \ref{fig:RL-CFE-generation}. This two-step process begins by passing an instance through a data transformation step, after which it is fed into a pretrained reinforcement learning policy to produce the CFE.

The proposed method is compared with a baseline approach, class-blind RL, which is identical to class-aware RL except that it excludes the instance's current predicted class from the state representation. This means that the RL agent doesn't have access to the model's prediction while it attempts to reach a state which is a counterfactual explanation for the provided instance. This serves as a strong baseline, as it isolates the effect of including class-awareness in the state representation, allowing for a clear assessment of its impact on the agent's performance. 
For evaluation, we tracked the convergence of rewards and episode length throughout the reinforcement learning training process. We also assessed the validity, sparsity, and proximity scores of the CFEs generated using our proposed state representation in comparison to the baseline. Additionally, we examined the impact of the instance's class on action selection by applying LIME \cite{Ribeiro2016"Why} and SHAP \cite{Lundberg2017Unified} methods to identify the key features in state representation influencing the RL model. Finally, we compared class-aware RL with two benchmark CFE approaches, DiCE \cite{mothilal2020explaining} and ReLAX \cite{Chen2022ReLAX:}.

The results demonstrate that class-aware RL offers a clear advantage over class-blind RL. Incorporating the instance's class accelerates policy convergence, enables higher rewards and shorter episode lengths, and generates significantly more valid CFEs. Additionally, the class-based feature consistently ranked among the top three state representation features influencing the learned policy's action selection.

The rest of the paper is structured as follows: Section \ref{lit-review} reviews related work, while Section \ref{method} outlines the problem formulation and details of the proposed method. Section \ref{experiment} describes the experimental setup, and Section \ref{result} presents the results. Finally, Section \ref{conclusion} concludes the paper.

\section{Literature Review}	
\label{lit-review}

LIME \cite{Ribeiro2016"Why}, SHAP \cite{Lundberg2017Unified}, and CFE \cite{wachter2017} are widely-used techniques for making black-box machine learning models explainable. These techniques have been used in different domains such as healthcare, finance, transportation, legal and education \cite{mathew2025recent, budhkar2025demystifying, khosraviExplainableArtificialIntelligence2022, kinjo2026fair}. The effectiveness of these techniques have been demonstrated with tabular \cite{Molnar2022Interpretable}, image \cite{gupta2025explainable}, and text \cite{alyoubi2025interpretable} for a variety of problems such as supervising learning (classification, regression) \cite{Molnar2022Interpretable} and unsupervised learning (clustering and recommendation) \cite{louhichiShapleyValuesExplaining2023}.

In the realm of CFE, a range of different approaches have been proposed. \cite{brughmans2024nice} categorizes CFE into four families: nearest unlike neighbors, generative methods, genetic algorithms, and SAT (Satisfiability) problems. Notably, \cite{wachter2017} minimizes a loss function to obtain counterfactual instances, considering the model's output, desired output, and instance distances. Another approach by \cite{Wexler2020What-If} identifies the nearest unlike instance, enhanced by the NICE algorithm using a breadth-first search (BFS) for sparser CFE \cite{brughmans2024nice}. Similar to NICE, \cite{crupi_counterfactual_2024} also relies on nearest unlike instance to determine a CFE. However, instead of feature space, the nearest unlike instance is determined in the latent space supported by structural causal models. TABCF is another method that uses latent space to generate CFE. It used transformer-based variational autoencoder to learn the latent space for tabular data \cite{panagiotou_tabcf_2024}. \cite{wiratunga2021discern} uses SHAP and LIME explainers for relevance-based CFE, while \cite{mothilal2020explaining} optimizes loss function via gradient descent, considering user context and constraints. \cite{Poyiadzi2020FACE:} emphasizes connectedness and density for CFE, \cite{Schleich2021GeCo:} introduces GeCo using a genetic algorithm for real-time CFE, and \cite{Rasouli2022CARE:} introduces CARE for coherent CFEs with user preferences.
Recent work has also explored genetic algorithms with adaptive feature weighting to improve sparsity and actionability in CFEs \cite{aljalaud2024counterfactual}. 
To tackle robustness issues after retraining of models, \cite{jiang_provably_2024} proposed PROPLACE to generate robust and plausible CFE.

Several studies delve into the generation of CFEs through RL. \cite{Li2017Understanding} employs a RL model in sentiment analysis, extracting a minimal set of tokens to alter a sentence's sentiment. For drug-target interaction models in XAI, \cite{Nguyen2021Counterfactual} introduces the MACDA framework, employing multi-agent RL to generate CFEs for drug-protein complexes. \cite{Numeroso2020Explaining} presents MEG, generating informative CFEs as structurally similar compounds with diverse predicted properties. \cite{Ezzeddine2023SAC-FACT:} introduces SAC-FACT, a RL-based system training individual soft actor critic models for each data point to generate corresponding CFE. \cite{Samoilescu2021Model-agnostic} proposes a deep reinforcement learning (DRL) approach generating batches of counterfactual instances in a single forward pass by modifying latent representations. \cite{Chen2022ReLAX:} introduces ReLAX, a model-agnostic algorithm crafting optimal counterfactuals using hybrid discrete-continuous actions and a curiosity explorer module. \cite{verma2022amortized} employs an amortized approach to train a RL model for faster CFE generation, considering sparsity, data manifold closeness, and causal relations.

In the literature, assessing CFE's quality involves ensuring validity \cite{mothilal2020explaining}. Desirable attributes, according to \cite{verma_counterfactual_2024}, include actionability, sparsity, adherence to data distribution, and causality. 
Discriminative power is another desirable property in CFE, but it can be difficult to measure due to its subjectivity \cite{guidotti2024counterfactual}. 
\cite{wachter2017,mothilal2020explaining} employ Manhattan distance normalized by inverse median absolute deviation, \cite{Poyiadzi2020FACE:} uses L2-norm, and \cite{Dandl2020Multi-Objective} applies the Gower distance function, treating numerical and categorical features differently. Proximity to the training data manifold enhances CFE reliability; \cite{Dandl2020Multi-Objective} estimates similarity with k-nearest neighbor, while \cite{brughmans2024nice} employs autoencoder's reconstruction loss, terming it plausibility.

None of the previous RL-based CFE studies have incorporated an instance?s predicted class into the state representation. This work fills that gap, showing that class-awareness enhances RL policy learning and improves the quality of generated CFEs.

\section{Methodology}
\label{method}

\begin{figure}[htbp]
	\centering
	\includegraphics[width=0.9\linewidth]{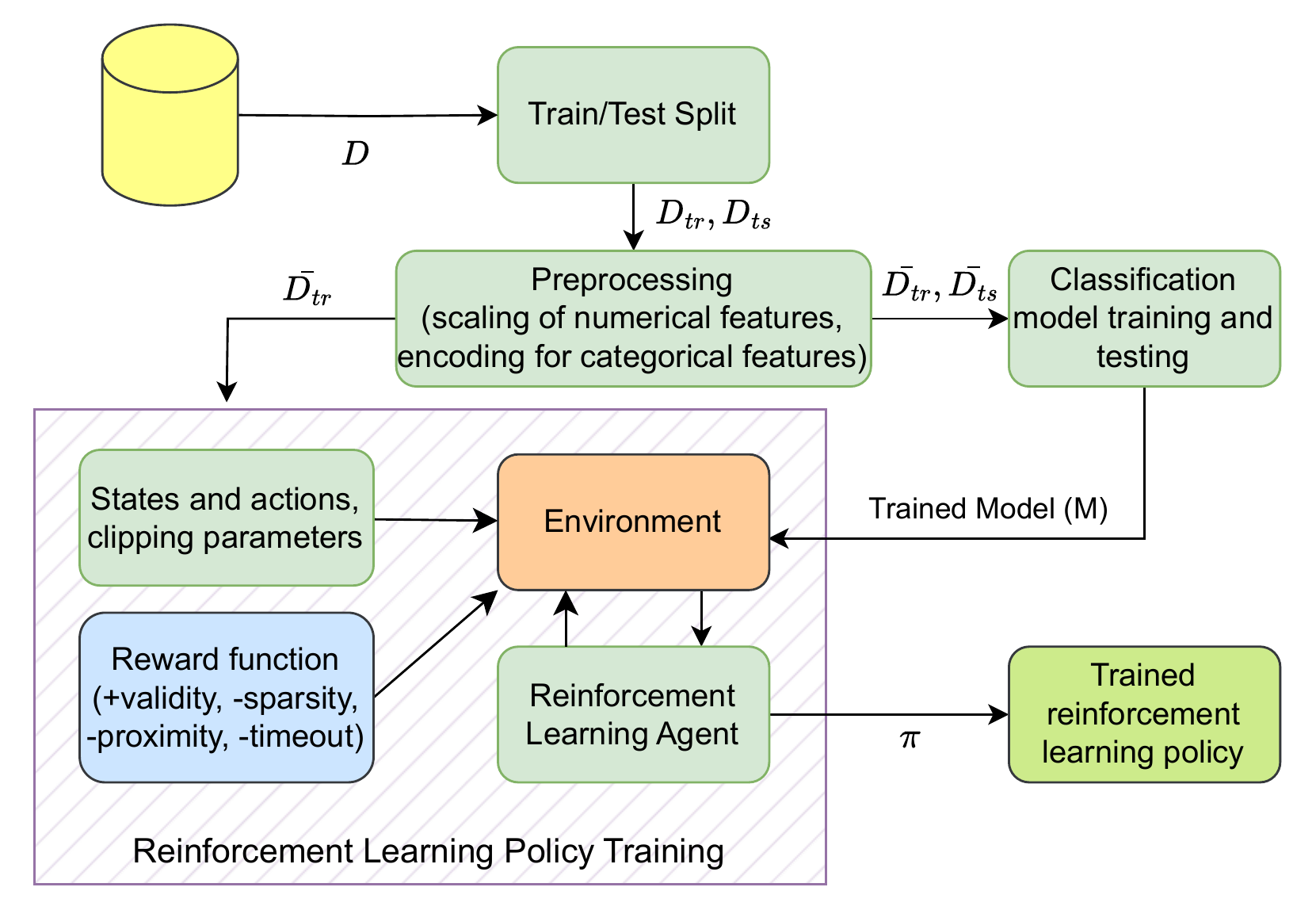}
	\caption{A workflow illustrating the process of end-to-end reinforcement learning policy training, including key steps such as data split, model training, and environment creation.}
	\label{fig:RL-CFE-training}
\end{figure}

This section details our methodology for incorporating the instance's class into the RL state representation for CFE generation. 
Figure \ref{fig:RL-CFE-training} shows the detailed process of generating CFEs using RL. The methodology consists of several components: a raw dataset, pre-processing module, supervised learning model, reinforcement learning model, and environment. 
The process begins by loading the raw dataset and splitting it into training and testing sets.
Formally, we assume a dataset, $D=\{X,Y\}$, where $X$ are the instances and $Y$ are the labels on the instances. $x \in X$ has $n$ features, $f_0, f_1, f_2, ..., f_n$. Furthermore, we assume a binary class problem, such that for a class label $y\in Y$, $y \in [0,1]$. In order to align with this objective, multi-class problems as posed in the penicillin dataset \cite{goldrickDevelopmentIndustrialscaleFedbatch2015a} (different types of faults + no fault)  are approached by employing a one versus all strategy (fault + no fault).

To begin, the dataset $D$ is split into training and testing sets in a 70:30 ratio. The purpose of the training data, $D_{tr}$, is to facilitate the training process for both the supervised learning model and RL policy. Conversely, the test data, $D_{ts}$, functions as a mean to assess and evaluate the performance of both models mentioned earlier.

The next step involves pre-processing the input features. 
A transformation, $Tr(f_n)$, is applied to the features in $x$.
The attributes in variable $x$ can take on discrete or continuous values. For the discrete features, a label-encoding technique is adopted, which assigns numerical codes ranging from 0 to ($l$-1), where $l$ represents the total number of distinct values present within each discrete feature. In order to process the continuous features, a min-max scaling technique is applied. By default, the resolution of the scaled features is set at 0.01 and the scaler values are rounded to two decimal place. It is important to highlight that depending on the specific problem at hand, one can adjust the resolution of numerical values as needed. The selection of resolution is a trade-off between faster policy learning versus proximity to the actual instance. The pre-processed data trains both supervised learning and reinforcement learning models. The same set of transformations is applied to both $D_{tr}$ and $D_{ts}$, resulting in transformed datasets $\bar{D_{tr}}$ and $\bar{D_{ts}}$.

A predictive model, $M$, is learned using a supervised learning algorithm on $\bar{D_{tr}}$ which serves as a black box model. We assume any supervised learning algorithm can be used e.g. Artificial Neural Network, Support Vector Machine, Deep Learning, etc. Our objective is to provide an explanation for the prediction $\bar{y}$ given by $M$ against an instance $x$. 
The explanation is given as a counterfactual instance $\bar{x}\neq x$ which has $n$ features $-$ identical to $x$ $-$ and an opposite class label $\hat{\bar{y}}$. The subsequent steps involve training a reinforcement learning policy, $\pi$, in a customized environment to generate a counterfactual instance $\bar{x}$, with details provided in the following subsections. The trained policy is then used to generate CFEs, as illustrated in Figure  \ref{fig:RL-CFE-generation}.

\subsection{Reinforcement Learning Formulation}
The underlying framework of our counterfactual instance generation process is Markov Decision Processes (MDP). Our formulation has four main components states, actions, transition function, and reward function, such that $MDP=\{S,A,T,R\}$. Additionally, we model it as an episodic task.

\begin{figure}[htbp]
	\centering
	\includegraphics[width=0.8\linewidth]{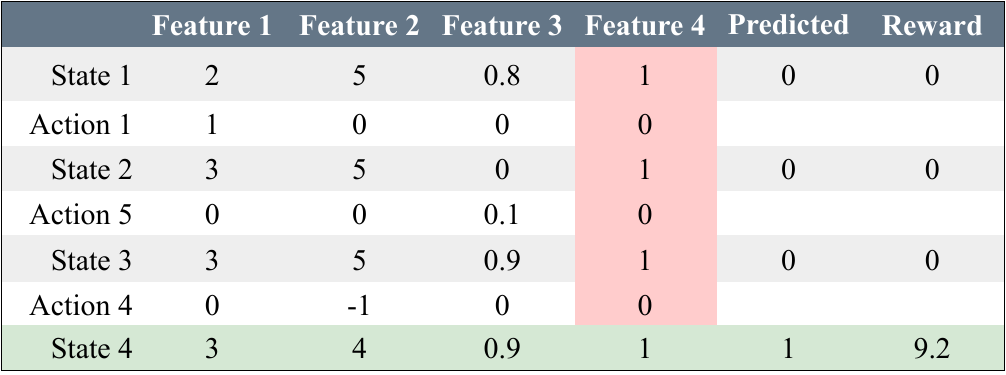}
	\caption{A simple example to show the representation of states, actions, transitions, and rewards. States are represented in odd-numbered rows, while actions are represented in even-numbered rows. Starting from row 1, taking the action in the subsequent row transitions the state to row 3. This process continues until a CFE is obtained, identified by a flipped predicted column. At this juncture, the reward is computed. Additionally, note that feature 4 is non-actionable and remains unchanged.}
	\label{fig:state-transitions-with-actions}
\end{figure}

\begin{itemize}
	\item \textbf{Episode:} The start state of an episode is derived using the instance for which a counterfactual is produced. 
	At the start of each training episode, a random instance from the $\bar{D_{tr}}$ is chosen by the environment for CFE generation, serving as the initial state. This sampling allows for repetition, meaning the same instance may be processed multiple times. For policy testing, a state representation based on a test instance is used as the starting state. The episode continues until the CFE is obtained or the maximum number of steps is reached. The training duration is influenced by the "number of steps to train" parameter, making it highly improbable for the RL model to encounter all training samples, particularly when dealing with larger datasets.
	
	\item \textbf{State:} A state $s\in S$ is represented by $n$ feature values and the class of the instance under observation. Moreover, the total number of states, $S$, is countable. As mentioned previously, these are transformed features that have undergone one of the transformations discussed above. 
	
	\item 	\textbf{Action:} A finite action set $A$ is assumed with the length equal to twice the number of features. Two actions for each feature correspond to a positive and a negative change respectively. 
	If a feature is immutable, its corresponding action is omitted from the action set $A$. However, it remains in the state representation, as it may provide useful information for learning. For instance, the class of the instance is always an immutable feature. In some cases, a feature can only be modified in one direction, which will result in further reduction of the action set by one action. 
	Selecting an action means selecting the feature to change its value. Therefore, an action changes a feature's value. After an action is selected by an action selection method, the selected feature's value is modified according to a fixed step-size. 
	For all datasets, categorical features were modified using a step size of $\pm$1 (due to integer encoding). For scalar features, we used $\pm$0.01 for Forest Cover, German Credit, Adult Income, and Heart Disease; and $\pm$0.1 for the rest, based on preliminary experimentation.
	An action modifies only a single dimension of the state unless it has dependency on another dimension. For example, in the case of one-hot encoded features, if one dimension belonging to a feature is set to 1, other dimensions belonging to the same feature must be set to 0. Note, this type action modification applied to only the Forest Cover dataset. Lastly, feature values are subject to clipping should a feature value extend beyond its acceptable range.

	A crucial aspect of our methodology involves how the actions are executed. The modification of the state is achieved through an additive process, where a vector representing an action is added to the existing state vector. Figure \ref{fig:state-transitions-with-actions} illustrates an example of state transitions resulting from the application of additive actions. 
	During training, actions are sampled from a stochastic policy learned by Proximal Policy Optimization (PPO), which naturally balances exploration and exploitation. During testing, actions are selected deterministically by choosing the most probable action from the learned policy.

	\item \textbf{Transition function:} We assume a deterministic transition function. After every step, the new state is the same as the old state except from the feature whose value is modified by the previous action.
	\item \textbf{Reward function:} We assign terminal rewards based on three criterion. First, if the class has been flipped then a positive reward (+10), $r_1$, is assigned. Second, a negative reward is assigned based on sparsity which measures the number of features modified (i.e., more features being changed will result in a higher negative reward). Third, a negative reward, $r_3$, is assigned based on proximity which the distance between original instance and the counterfactual instance (i.e., the greater the distance, the higher the negative reward). For proximity, we utilized the cosine distance due to its inherent constraint within the range of 0 to 1. Therefore, the terminal reward is the sum, $r_1 + r_2 + r_3$. No intermediate rewards are assigned in an episode. Moreover, if the number of steps in an episode exceeds the maximum allowed, a small negative reward (-1) is assigned.
\end{itemize}

Figure \ref{fig:state-transitions-with-actions} shows the states before and after an action, as well as the rewards assigned over three steps in an example with four features, including three mutable features. Note that six actions are possible: to increase or decrease the values of Features 1, 2, and 3.
Initially, Action 1 is undertaken by the RL agent resulting in a positive modification to Feature 1. Following this action, all aspects of the state remain unchanged except for Feature 1 which increases by one unit and causes a transition from State 1 to State 2 within the environment. Subsequently, after each action has been implemented, the change in class of the state determines whether CFE has been obtained or not. If an explanation is generated, it signals that episode completion should occur and rewards can be calculated accordingly. Otherwise, continuing onwards will prompt further actions to be taken by RL agent. Since the class was not flipped by Action 1, another action is taken i.e. Action 5, which modifies the state of the environment from State 2 to State 3. Again, the CFE is not obtained and therefore RL agent will continue to take another action. This time, Action 4 is taken which modified Feature 2 in the negative direction transitioning the environment into State 4. State 4 eventually flips the class and a CFE is obtained. Reward is also computed at this point.

\subsection{Reinforcement Learning Method}
The RL agent learns a policy using the Proximal Policy Optimization (PPO) algorithm. 
Fundamentally, PPO is an on-policy, actor-critic and policy gradient method which searches for an optimal policy in the policy space by applying gradient decent on the objective function (OF). PPO  uses a clipped OF which disallows large updates in a gradient decent step. For this, it uses a clipping parameter that limits the change in the policy during each update. For further details about PPO, we refer the reader to the original paper \cite{schulman2017proximal}.

\subsection{Baseline method}
Our objective is to examine whether awareness of the class label helps the RL agent identify the best actions.
For comparison, we use a class-blind approach as a baseline, which follows the same method as discussed above but excludes the class label from the state representations. This serves as a good baseline, as it allows us to isolate and evaluate the specific impact of the class label on the learner's performance. Note that excluding the class label aligns this approach closely with the FastAR method proposed by \cite{verma2022amortized}, but we focus solely on terminal rewards and omit causal constraints to isolate the impact of our proposed modifications.

\subsection{Environment Setup}
Setting up an environment is essential to our methodology. The environment provides an implementation of the components of an MDP, including states, actions, transitions, and rewards. Additionally, it defines the logic for episode initialization and termination. A separate environment is created for each dataset, using Python's open-source library, Gymnasium \cite{towers_gymnasium_2023}. This environment is used both to train the reinforcement learning policy and to apply the learned policy for CFE generation. Each environment requires the following inputs: the training dataset $\bar{D_{tr}}$, the test dataset $D_{ts}$, and the classification model $M$. In return, it provides functionality to initialize an episode, execute actions, and receive rewards along with information about episode termination. Additional parameters like the maximum number of steps and feature clipping values are defined in the environment.

\subsection{Evaluation using Convergence Graphs}
To evaluate the performance of the proposed method, we analyzed the convergence graphs for both the reward and the episode across all datasets, comparing them against the baseline approach. The convergence graphs were generated by running each experiment 10 times, ensuring that the results are statistically robust. At each time step, the convergence graphs display the mean values calculated from all the runs, providing a comprehensive view of how the reward and episode metrics evolve over time for both the proposed approach and the baseline. These graphs provide insights into the speed of reward convergence, the stability of the converged values (as indicated by variance), and the final convergence values.

\subsection{Evaluation using XAI-based Feature Importance Measures}

To assess the effectiveness of including the instance's class in the RL state representation, we also employed LIME \cite{Ribeiro2016"Why} and SHAP \cite{Lundberg2017Unified} to identify the most influential features affecting the action selection of the learned policy $\pi$. Given that the RL model operates on the current environmental state, we constructed a dataset of these states for analysis. First, we utilized the test set and generated CFEs for each instance using $\pi$. During the CFE generation process, we recorded all visited states. To manage computational complexity, we randomly sampled a smaller subset of states from the collected dataset. For each sampled state, we applied LIME and SHAP to determine the importance of individual features. This provides local feature importance, specific to the current state under observation. To obtain a broader perspective, we calculated the average feature importance across all samples. This provides insights into the overall significance of each feature in the RL decision-making process. By comparing the feature importance rankings with and without the included class, we evaluate the impact of our proposed modification on the RL agent's decision-making behavior.

\subsection{Comparison with Benchmark CFE Methods}
We compared class-aware RL with two established CFE methods: DiCE \cite{mothilal2020explaining}, a gradient-based approach, and ReLAX \cite{Chen2022ReLAX:}, an RL-based approach, both of which have shown superior performance over traditional neural network? and optimization-based techniques.


\section{Experimentation}
\label{experiment}

The datasets utilized in this study are cited in Table \ref{table:datasets}. Few of the chosen datasets were previously employed for generating CFE using RL. In addition to these, we have incorporated penicillin, forest cover, and heart disease datasets with larger dimensions to assess the scalability of our approach. To the best of our knowledge, penicillin and heart disease datasets have not been used in previous literature for generating CFEs. The accuracies of the classification model on the test split for each dataset are specified in the 'Accuracy' column in Table \ref{table:datasets}.

\begin{table}[h]
	\centering
	\label{table:datasets}
	\footnotesize 
	\setlength{\tabcolsep}{4pt} 
	\begin{tabular}{@{}l l l r@{}}
		\toprule
		\textbf{Dataset} & \textbf{Dimensions} & \textbf{Features} & \textbf{Accuracy} \\
		\midrule
		Breast Cancer   & 569 $\times$ 31      & Numerical & $0.959 \pm 0.015$ \\
		German Credit   & 1,000 $\times$ 21    & Numerical, Categorical & $0.757 \pm 0.010$ \\
		Adult Income    & 30,725 $\times$ 12   & Numerical, Categorical & $0.729 \pm 0.002$ \\
		Default Credit  & 30,000 $\times$ 24  & Numerical, Categorical & $0.820 \pm 0.002$ \\
		Penicillin      & 233,595 $\times$ 36  & Numerical & $0.600 \pm 0.003$ \\
		Heart Disease   & 319,795 $\times$ 18  & Numerical, Categorical & $0.730 \pm 0.002$ \\
		Forest Cover    & 581,012 $\times$ 55  & Numerical, One-hot encoded & $0.786 \pm 0.001$ \\
		\bottomrule
	\end{tabular}
	\caption{Summary of datasets: Large (Forest Cover, Penicillin, Heart Disease); Medium (Default Credit, Adult Income); Small (German Credit, Breast Cancer).}
	
\end{table}

\begin{figure*}[!ht]
	\centering
	\includegraphics[width=\textwidth]{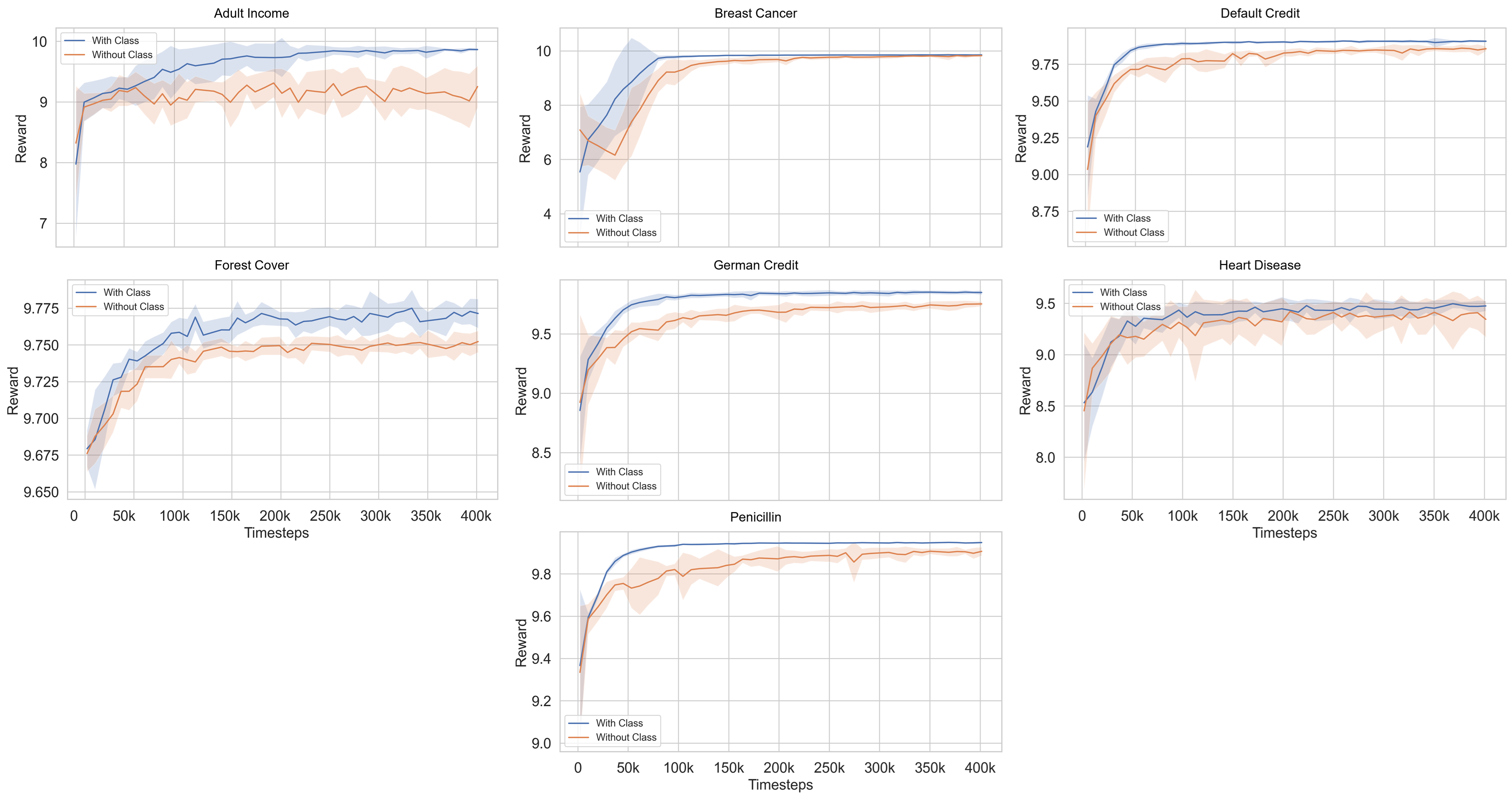}
	\caption{Evaluating training performance by comparing the mean reward (top) and episode length (bottom) for the class-aware (blue trace) and class-blind (orange trace) methods.}
	\label{fig:training_performance}
\end{figure*}

The experiments were implemented in Python using various libraries and frameworks, including pandas, numpy, sklearn, xgboost, gymnasium \cite{towers_gymnasium_2023} for creating environments, and stable-baselines3 \cite{stable-baselines3} for RL model. We utilized lime\footnote{https://lime-ml.readthedocs.io/en/latest/} and shap\footnote{https://shap.readthedocs.io/en/latest/} libraries for the implementation of LIME and SHAP, respectively. 

During the experimentation, we observed that XGBoost, with 200 estimators and a learning rate of 0.05, outperformed the other models without extensive fine-tuning efforts. Consequently, it was chosen as the predictive model across all our experiments. Other hyper-parameters such as max depth and positive weight were fine-tuned to obtain better results on specific datasets. For our RL model, PPO was chosen based on better results during the initial phase of the experimentation.
The PPO algorithm, as implemented in the stable-baselines3 library \cite{stable-baselines3}, was utilized with its default hyperparameter settings.

A conventional desktop computer was utilized for our study, equipped with an Intel Core i5-3470S CPU and 16 GB of RAM. However, it did not have a CUDA-supported GPU. The operating system running on the machine was Windows 10 64-bit. Python version 3.10.5 served as the primary programming language in our analysis.

For the assessment of the produced CFE, we assessed their validity, sparsity, proximity (using cosine distance), episode length, and terminal reward. To ensure fairness in comparison across different scenarios, we normalized both sparsity and proximity metrics. Note that for the class-aware RL method, the predicted class-based feature was excluded when computing these measures. As for the other metrics like validity and episode length, they were reported as originally obtained since these metrics can differ depending on the specific problem under consideration.

\begin{sidewaystable}
\setlength{\tabcolsep}{12pt}
	\centering
	\renewcommand{\arraystretch}{1} 
	\begin{tabular}{rcccccc}
		\hline
		\textbf{}               & \multicolumn{3}{c}{\textbf{Without Class Representation}}        & \multicolumn{3}{c}{\textbf{With Class Representation}}               \\
		\textbf{Dataset}        & \textbf{Validity} & \textbf{Sparsity}   & \textbf{Proximity}     & \textbf{Validity}    & \textbf{Sparsity}    & \textbf{Proximity}     \\ \hline
		\textbf{Breast Cancer}  & 0.95 $\pm$ 0.03       & 0.12 $\pm$ 0.01         & 0.04 $\pm$ 0.01            & \textbf{0.99 $\pm$ 0.01} & \textbf{0.11 $\pm$ 0.01} & 0.04 $\pm$ 0.01            \\ 
		\textbf{German Credit}  & 0.91 $\pm$ 0.03       & 0.14 $\pm$ 0.01         & 0.08 $\pm$ 0.01            & \textbf{0.97 $\pm$ 0.01} & \textbf{0.1 $\pm$ 0.01}  & \textbf{0.05 $\pm$ 0.01}   \\
		\textbf{Adult Income}   & 0.88 $\pm$ 0.03       & \textbf{0.1 $\pm$ 0.01} & \textbf{0.002 $\pm$ 0.003} & \textbf{0.95 $\pm$ 0.02} & 0.11 $\pm$ 0.01          & 0.01 $\pm$ 0.0             \\
		\textbf{Default Credit} & 0.91 $\pm$ 0.03       & 0.07 $\pm$ 0.0          & 0.027 $\pm$ 0.003          & \textbf{0.99 $\pm$ 0.01} & \textbf{0.06 $\pm$ 0.0}  & \textbf{0.02 $\pm$ 0.001}  \\ 
		\textbf{Penicillin}     & 0.94 $\pm$ 0.03       & 0.06 $\pm$ 0.0          & 0.011 $\pm$ 0.004          & \textbf{0.99 $\pm$ 0.0}  & \textbf{0.05 $\pm$ 0.0}  & \textbf{0.002 $\pm$ 0.0}   \\ 
		\textbf{Heart Disease}  & 0.92 $\pm$ 0.06       & 0.2 $\pm$ 0.01          & \textbf{0.302 $\pm$ 0.014} & \textbf{0.97 $\pm$ 0.03} & \textbf{0.19 $\pm$ 0.0}  & 0.303 $\pm$ 0.01           \\ 
		\textbf{Forest Cover}   & 0.99 $\pm$ 0.0        & 0.04 $\pm$ 0.0          & 0.208 $\pm$ 0.004          & \textbf{1.0 $\pm$ 0.0}   & 0.04 $\pm$ 0.0           & \textbf{0.184 $\pm$ 0.004} \\ \hline
	\end{tabular}%
	\caption{Evaluating test performance across all the datasets by comparing validaity, sparsity, and proximity scores of the generated CFE with and without the instance's class inclusion in the RL state.}
	\label{table:test_performance}
\end{sidewaystable}

\begin{figure*}[!ht]
	\centering
	\includegraphics[width=0.9\textwidth]{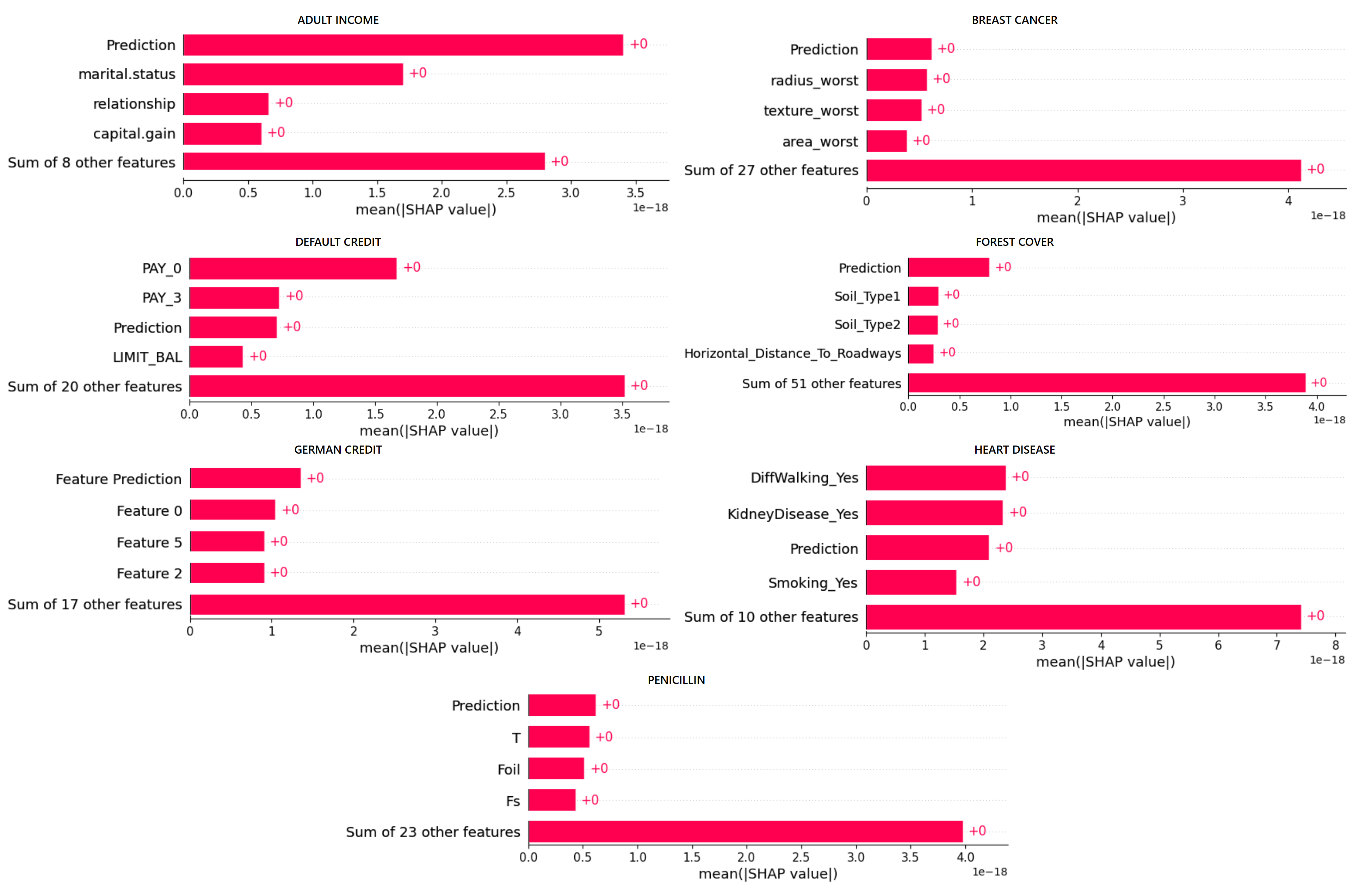}
	\caption{Top 5 most important features for RL action selection across all datasets, as determined using SHAP.}
	\label{fig:shap_feature_importance}
\end{figure*}

\begin{figure*}[!ht]
	\centering
	\includegraphics[width=0.9\textwidth]{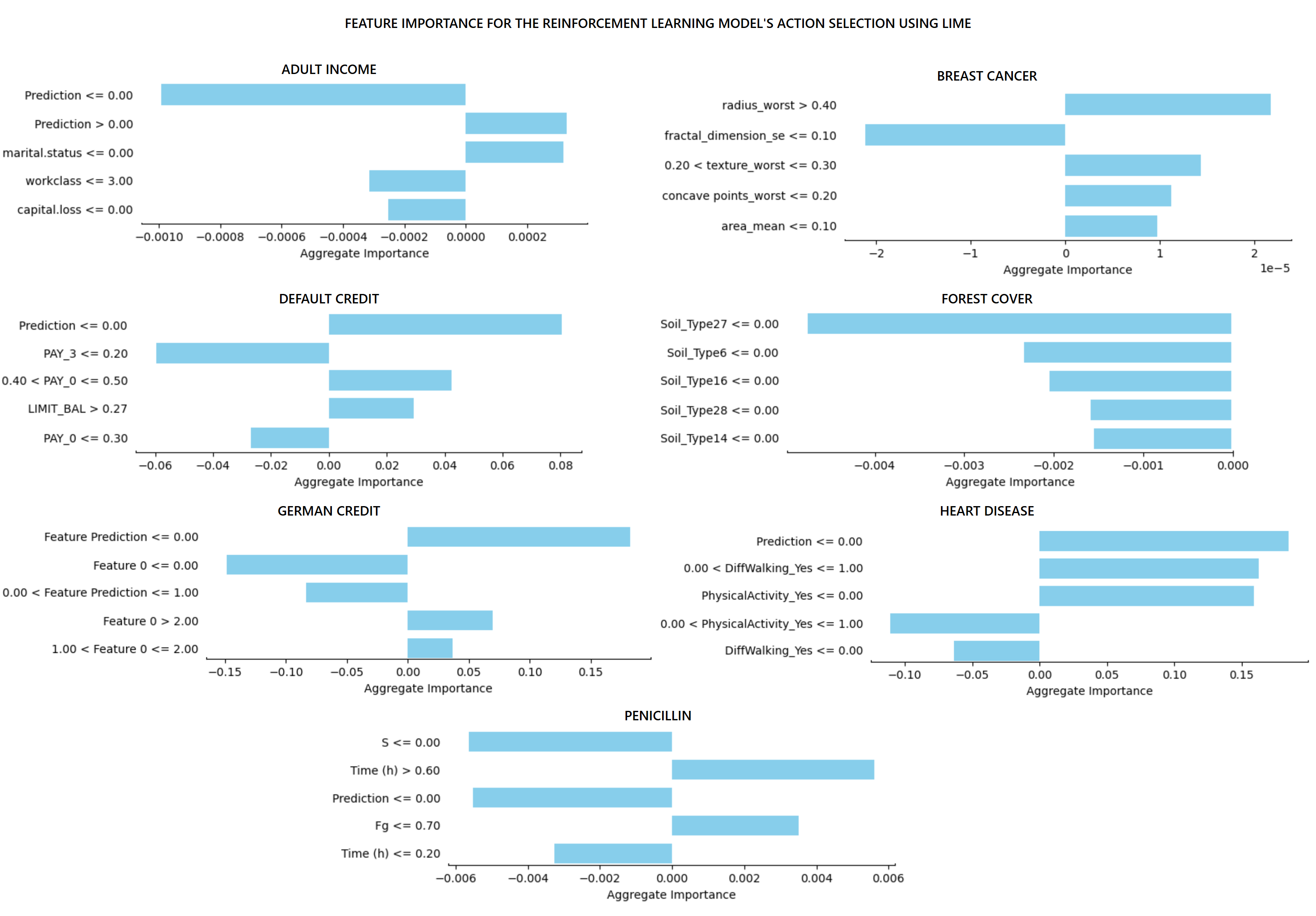}
	\caption{Top 5 most important features for RL action selection across all datasets, as determined using LIME.}
	\label{fig:lime_feature_importance}
\end{figure*}

\section{Results and Discussion}
\label{result}

This section is organized as follows: First, we discuss the training results in terms of the rewards and episode length convergence graphs. Next, we shift our focus to test performance, comparing the class-aware approach against the baseline and examining the sparsity, proximity, and validity of the generated CFEs. Then, we explore the impact of including the instance?s class by reviewing feature importance results obtained through LIME and SHAP. Finally, we compare our method with two established CFE techniques.

\subsection{Training Performance against baseline}

We tested the effectiveness of including the instance class for all the datasets by assessing its impact on episode reward and episode length. As mentioned before, the experiments were run 10 times for statistical robustness. The runs with and without the instance's class were trained for the same number of time steps. 

The training performance results for all the datasets are shown in Figure \ref{fig:training_performance}. It can be observed across all the datasets that inclusion of the instance's class in the state resulted in higher episode reward and smaller episode length. However, in the case of the Breast Cancer dataset, the rewards and episode lengths converged to nearly the same values for both the class-aware and class-blind methods. Nevertheless, the class-aware method demonstrated better performance in terms of early convergence. In addition, the class-aware method achieved  relatively faster convergence in three other datasets: Default Credit, German Credit, and Penicillin. Only in the case of Adult Income did the class-aware method converge more slowly than the class-blind method. However, this slower convergence resulted in much better final values for rewards and episode lengths, with almost zero variance.
The variance for reward and episode length was almost the same at the start of the training in all cases. However, at the end of the training, lower variance was observed in the case of class-aware method, indicating superior convergence performance. The episode reward and episode length plateaued at a worse value for the case when instance's class was not included. 

Overall, it can be observed that inclusion of the instance's class positively impacts the search for better CFE. In each case, we observed that the class-aware RL method yielded better reward and episode length values, and converged faster compared to the class-blind RL method. However, the difference in performance of the two algorithms was more prominent in some cases.

\subsection{Test Performance against baseline}

Table \ref{table:test_performance} highlights the test performance of class-aware RL against the baseline (i.e., class-blind RL). Starting with validity, RL performance with the class label is a clear winner with as much as 8\% absolute difference for Default Credit dataset and at least 1\% absolute difference for the Forest Cover dataset. 
Moreover, class-aware RL significantly improved the validity scores compared to the class-blind RL (p=0.0071, two-tailed t-test assuming unequal variances).
For sparsity, CFE generated with class-aware RL are sparser for 5 out of the 7 datasets, less sparse for 1 dataset (Adult Income) and same for 1 dataset (Forest Cover) against the baseline. When the obtained CFE is sparser, the sparsity improves as much as 0.04 points (German Credit) and otherwise is at least 0.01 point better. When sparsity is lower, the difference is 0.01 point. Finally, the proximity score for CFE generated with state modification is better for 4 out of the 7 datasets, worse for 2 datasets (Adult Income and Heart Disease), and same for 1 dataset (Breast Cancer). Where proximity score is better, the score improves as much as 0.058 points (Penicillin) and at least better by 0.02 points for the other datasets. 

One reason we observe a dip in scores for some datasets across sparsity and proximity metrics could be due to the higher reward assigned for validity (10) compared to the rewards assigned for sparsity (1 - sparsity score) and proximity (1 - proximity score). Although both class-aware RL and class-blind RL used the same reward structure, the class-aware RL approach demonstrated superior ability in leveraging these rewards to generate a higher number of valid CFEs compared to the baseline. As a result, while the validity scores improved, the sparsity and proximity scores dipped for some datasets. Further investigation into this trade-off between validity and other metrics is left for future research.

\subsection{Significance of instance's class in state representation}
To further validate our findings that the inclusion of an instance's class is indeed effective, we employed XAI techniques to determine whether the class ranks among the top features influencing the RL model's decision-making when selecting an action to generate counterfactual explanations (CFEs). For this analysis, we used both SHAP and LIME to obtain feature importance scores.

\subsubsection{Feature importance using SHAP}
Figure \ref{fig:shap_feature_importance} illustrates the feature importance for all datasets, showing the top 4 features along with a combined category for the remaining features. In the figure, the instance's class is labeled as 'Prediction'. For 5 out of the 7 datasets, 'Prediction' emerges as the most influential feature in the decision-making process for selecting an action. For the remaining datasets, it consistently ranks among the top 3 features. The magnitude shown represents the discriminative power of each feature. In some cases, such as Adult Income and Forest Cover, the difference in importance between 'Prediction' and the next most important feature is significant, while in other cases, the difference is smaller, highlighting the impact of including the instance's class.

\subsubsection{Feature importance using LIME}
Figure \ref{fig:lime_feature_importance} shows the feature importance rankings (top 5) for all datasets as determined by LIME. In the figure, the instance's class is labeled as 'Prediction.' Compared to SHAP, LIME's results are not as consistent. For 4 out of the 7 datasets, 'Prediction' emerged as the most influential feature, while for 1 dataset (Penicillin), it ranked among the top 3 features. However, for the remaining two datasets, 'Prediction' did not appear in the top 5 features.

Given that LIME and SHAP are distinct techniques with different underlying methodologies, their results may not always align. In our case, there was partial overlap, as both methods identified the instance's class as the most influential feature for two datasets (Adult Income and German Credit).

\subsection{Comparative Analysis}
Table \ref{tab:method-comparison-final} presents the results of two existing CFE techniques, DiCE and ReLAX, which we compare with our proposed method (Table \ref{table:test_performance}).
To ensure fair comparison, results for DiCE on the Forest Cover dataset were excluded since it requires specifying the opposite class in this multi-class setting. 
We also tested three DiCE variants (random, genetic and kd-tree) but only reported the random variant as it performed better overall than the other variants. The standard errors of the validity values were zero in all cases. Note that in cases where no value is highlighted, class-aware RL achieved better or the same results.

No single approach dominates across all datasets. 
Despite consistently achieving approximately $99\%$ validity, our method exhibited slightly lower performance compared to other techniques in this aspect. 
In terms of sparsity, our method outperformed other algorithms, obtaining the best results for datasets such as adult income, penicillin, forest cover, and default credit. In datasets beyond these, our method's performance remained competitive with other techniques. Additionally, our method exhibited commendable performance in proximity, achieving joint-best results for adult income and marginally better results for the penicillin dataset.

\begin{table}[!htbp]
	\centering
	\small 
	\setlength{\tabcolsep}{2pt} 
	
	\caption{Results for benchmark CFE generation methods. The highlighted values indicate better results compared to class-aware RL.}
	\label{tab:method-comparison-final}
	
	\begin{tabular}{@{} l l l l l l @{}}
		\toprule
		\textbf{Dataset} & \textbf{Method} & \textbf{Size} & \textbf{Validity} & \textbf{Sparsity} & \textbf{Proximity} \\
		\midrule
		
		\multirow{2}{*}{\makecell{Breast\\Cancer}}
		& DiCE & 171 & $\mathbf{\mathbf{1.00}}$ & $0.12 \pm 0.01$ & $0.15 \pm 0.02$ \\
		& ReLAX & 171 & $\mathbf{\mathbf{1.00}}$ & $\mathbf{0.06 \pm 0.00}$ & $0.09 \pm 0.00$ \\
		\cmidrule(l){2-6}
		
		\multirow{2}{*}{\makecell{German\\Credit}}
		& DiCE & 300 & $\mathbf{\mathbf{1.00}}$ & $0.11 \pm 0.00$ & $0.09 \pm 0.00$ \\
		& ReLAX & 300 & $\mathbf{\mathbf{1.00}}$ & $\mathbf{0.09 \pm 0.00}$ & $\mathbf{0.04 \pm 0.00}$ \\
		\cmidrule(l){2-6}
		
		\multirow{2}{*}{\makecell{Adult\\Income}}
		& DiCE & 9218 & $\mathbf{\mathbf{1.00}}$ & $0.17 \pm 0.00$ & $0.09 \pm 0.00$ \\
		& ReLAX & 9218 & $\mathbf{\mathbf{1.00}}$ & $0.16 \pm 0.00$ & $0.09 \pm 0.00$ \\
		\cmidrule(l){2-6}
		
		\multirow{2}{*}{\makecell{Default\\Credit}}
		& DiCE & 5000 & $\mathbf{\mathbf{1.00}}$ & $0.08 \pm 0.00$ & $0.15 \pm 0.01$ \\
		& ReLAX & 5000 & $\mathbf{\mathbf{1.00}}$ & $0.10 \pm 0.00$ & $0.097 \pm 0.000$ \\
		\cmidrule(l){2-6}
		
		\multirow{2}{*}{Penicillin}
		& DiCE & 5000 & $\mathbf{\mathbf{1.00}}$ & $0.07 \pm 0.00$ & $0.004 \pm 0.000$ \\
		& ReLAX & 70079 & $\mathbf{\mathbf{1.00}}$ & $0.07 \pm 0.00$ & $0.002 \pm 0.000$ \\
		\cmidrule(l){2-6}
		
		\multirow{2}{*}{\makecell{Heart\\Disease}}
		& DiCE & 5000 & $\mathbf{\mathbf{1.00}}$ & $\mathbf{0.14 \pm 0.00}$ & $0.130 \pm 0.000$ \\
		& ReLAX & 95939 & $\mathbf{\mathbf{1.00}}$ & $0.20 \pm 0.00$ & $\mathbf{0.129 \pm 0.000}$ \\
		\cmidrule(l){2-6}
		
		\multirow{2}{*}{\makecell{Forest\\Cover}}
		& DiCE & -- & -- & -- & -- \\
		& ReLAX & 174304 & $1.00$ & $0.13 \pm 0.00$ & $0.231 \pm 0.000$ \\
		\bottomrule
	\end{tabular}
\end{table}

\section{Conclusion}
\label{conclusion}

In this study, we analyzed the impact of including or excluding the instance's class as a feature in the state representation for RL to generate counterfactual explanations. 
A comparison of the two approaches was performed using seven datasets from diverse domains and varying in size, ranging from small to large.
Our work shows that for CFE generation via RL, the inclusion of instance's class in the state representation provides better training performance.
The experiments confirmed that during the training phase, class-aware RL method either converges to better values for rewards and episode lengths, achieves faster convergence than the class-blind RL method, or both.
Adding instance's class information to RL's states also enhances the quality of generated CFEs in terms of various important measures like validity, sparsity and proximity.
During the test phase, class-aware RL consistently generated significantly more valid CFEs across the datasets compared to the baseline.
Moreover, class-aware RL demonstrated superior performance in terms of sparsity and proximity across the majority of datasets. 
Our findings were further supported by feature importance results derived from two different explainability techniques, SHAP and LIME. The feature representing class instance information consistently ranked among the top three features according to SHAP scores across all datasets. However, for LIME scores, this feature ranked among the top three in all but two datasets.
Overall, our work demonstrates that including the instance's class into the state representation of RL for CFE generation provides performance gain during both training and testing phases. Additionally, our findings provide empirical evidence that the class-based feature is often among the most influential features, aiding the RL agent in generating CFEs effectively. 
Our work highlights the necessity for a deeper understanding of how instance's class information can influence CFE generation and decision-making of the RL agent.
Future research could expand this analysis to non-tabular data, including images or text, to investigate whether similar patterns emerge across these diverse domains. 
Furthermore, it is recommended to conduct a more in-depth exploration into optimizing reward functions for RL models. Lastly, assessing utility on downstream tasks remains an important direction for future work.

\backmatter

\bmhead{Supplementary information}

The code is provided to run the algorithms.

\bmhead{Data availability} The datasets analyzed in this study are publicly available. The Breast Cancer, German Credit, Adult Income, Default Credit, and Forest Cover datasets were obtained from the UCI Machine Learning Repository (https://archive.ics.uci.edu/). For the Adult Income dataset, a pre-processed (cleaned) version was utilized. The Heart Disease dataset was sourced from the CDC's Personal Key Indicators of Heart Disease via Kaggle (https://www.kaggle.com/datasets/kamilpytlak/personal-key-indicators-of-heart-disease). The Penicillin dataset was generated using the simulator available at the Industrial Penicillin Simulation website (http://www.industrialpenicillinsimulation.com/). All relevant data used during this study are included in the Supplementary Information files.


\bibliography{sn-bibliography}

\end{document}